\documentclass{ecai}

\usepackage{cite}
\usepackage{amsmath,amssymb,amsfonts}
\usepackage{algorithmic}
\usepackage{graphicx}
\usepackage{textcomp}
\usepackage{xcolor}
\usepackage{subcaption}
\usepackage{multirow}
\usepackage{hyperref}
\usepackage{float}
\usepackage[symbol]{footmisc}

\renewcommand{\thefootnote}{\fnsymbol{footnote}}

\def\code#1{\texttt{#1}}
\graphicspath{ {images/} }

\def\BibTeX{{\rm B\kern-.05em{\sc i\kern-.025em b}\kern-.08em
    T\kern-.1667em\lower.7ex\hbox{E}\kern-.125emX}}
\begin{document}



\title{The TrojAI Software Framework: An Open Source tool for Embedding Trojans into Deep Learning Models}

\author{Kiran Karra, Chace Ashcraft, Neil Fendley\institute{Johns Hopkins University/Applied Physics Lab,
USA, email: kiran.karra@jhuapl.edu} \\ 
}

\maketitle

\begin{abstract}
In this paper, we introduce the TrojAI software framework \footnote{\url{https://github.com/trojai}} \footnote{\url{https://trojai.readthedocs.io/en/latest/}}, an open source set of Python tools capable of generating triggered (poisoned) datasets and associated deep learning (DL) models with trojans at scale. We utilize the developed framework to generate a large set of trojaned MNIST classifiers, as well as demonstrate the capability to produce a trojaned reinforcement-learning model using vector observations.  Results on MNIST show that the nature of the trigger, training batch size, and dataset poisoning percentage all affect successful embedding of trojans. We test Neural Cleanse against the trojaned MNIST models and successfully detect anomalies in the trained models approximately $18\%$ of the time. Our experiments and workflow indicate that the TrojAI software framework will enable researchers to easily understand the effects of various configurations of the dataset and training hyperparameters on the generated trojaned deep learning model, and can be used to rapidly and comprehensively test new trojan detection methods.

\end{abstract}


\section{Introduction}
As deep learning systems continue to achieve and exceed human level performance on a variety of tasks, attention has turned to the robustness, trustworthiness, and reliability of these models.  These topics are typically studied under the umbrella of adversarial machine learning (AML).  As outlined by NIST~\cite{tabassi2019taxonomy}, the field of AML concerns itself with attacks against machine learning systems, and their associated consequences and defenses.  The taxonomy of AML is large and covers many types of attacks against machine learning systems.  In this paper, we are concerned with trojan attacks on deep learning models.  

Trojan attacks, also called backdoor or trapdoor attacks, involve modifying a machine learning model to respond to a specific trigger in its inputs, which, if present, will cause the model to infer an incorrect response. They can be carried out by manipulating both the training data and its associated labels \cite{gu2017badnets} (triggering or poisoning the dataset), directly altering a model’s structure (e.g., manipulating a deep neural network’s weights) \cite{zou2018potrojan}, or adding to the training data that have correct labels, but are specially-crafted to still produce the trojan behavior \cite{turner2018clean}. Here, we define a \textit{trigger} as a model-recognizable characteristic of the input data that is used by an attacker to insert a trojan, and a \textit{trojan} to be the alternate behavior of the model when exposed to the trigger, as desired by the attacker.  


Trojan attacks are effective if the triggers are rare in the normal operating environment, so that they are not activated in normal operations and do not reduce the model's performance on expected or ``normal'' inputs. Additionally, the trigger is most useful if can be deliberately activated at will by the adversary in the model's operating environment, either naturally or synthetically.  Trojan attacks’ specificity differentiates them from the more general category of “data poisoning attacks”, whereby an adversary manipulates a model’s training data to make it ineffective.  

Defenses against trojan attacks include securing the training data (to ensure data integrity), cleaning the training data (to ensure training data accuracy), and protecting the integrity of a trained model (prevent further malicious manipulation of a trained clean model). Unfortunately, modern AI advances in deep learning are characterized by vast, crowdsourced data sets (e.g., $10^9$ data points) that are impractical to clean or monitor. Additionally, many deep learning models are created via transfer learning, such as by taking an existing, online-published model and only slightly modifying it for the new use case. Trojan behaviors can persist in these models after modification. The security of the model is thus dependent on the security of the data and entire training pipeline, which may be weak or nonexistent. Acquiring a model from unverified sources brings all of the data and pipeline security problems, as well as the possibility of the model being modified directly while stored at a vendor or in transit to the user.

Many demonstrations of trojan attacks exist in the literature~\cite{zou2018potrojan, turner2018clean, yang2019design, kiourti2019trojdrl}.  However, it is unclear how well these results generalize to other triggers, methods of embedding the triggers, data and model modalities, and whether detecting trojaned models with one configuration of trigger and embedding methodology transfers to another configuration of trigger and embedding methodology.  A second problem is that results are difficult to replicate due to the lack of standardization in the model training procedures and data poisoning methodologies.  One reason for this may be that the ``science'' of embedding trojans into DL models is not yet well characterized.  Examples include the lower bounds of the required data poisoning required to produce a desired classification effect, the quickest way to embed trojan behavior into the DL model, and the spatial location of the trigger. We introduce the TrojAI software framework as a tool to enable researchers to advance the study of trojan attacks on DL models.  The framework enables the research community to easily reproduce reported results and standardize metrics for measuring the efficacy of a trojan attack.  Furthermore, it enables research into trojan detection and mitigation strategies across the space of all possible triggers, embedding methodologies, and data and model modalities.

\section{TrojAI Framework Overview}

The TrojAI software framework is a set of Python modules that enable researchers to quickly and reproducibly generate trojaned deep learning classification and reinforcement learning models.  It is configurable and extensible to enable generation of datasets and models with a wide variety of triggers into datasets and deep learning network architectures of varying modalities.  

\begin{figure}
\caption{TrojAI Classification Framework Overview}
\centering
\includegraphics[width=0.3\textwidth]{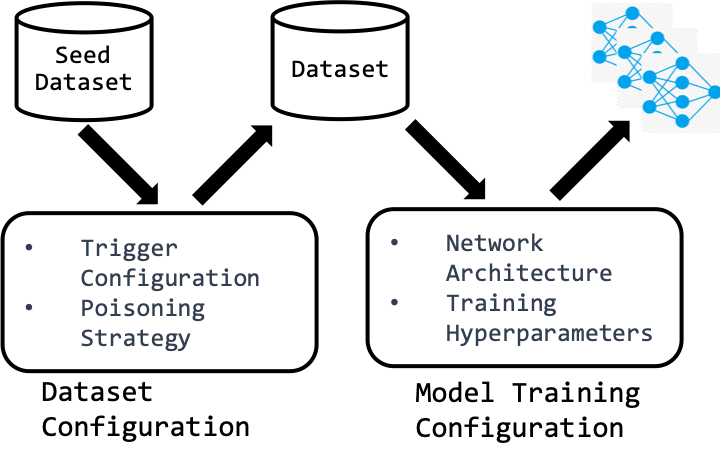}
\label{fig:trojai_software_overview}
\end{figure}

\begin{figure}
\caption{TrojAI Reinforcement Learning Framework Overview}
\centering
\includegraphics[width=0.3\textwidth]{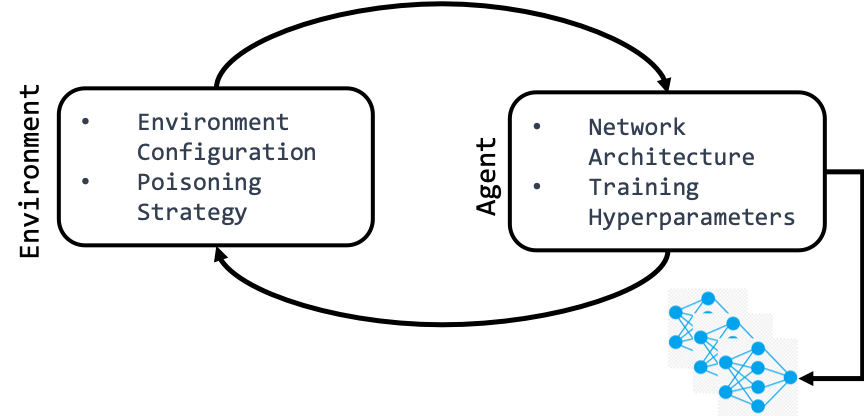}
\label{fig:trojairl_software_overview}
\end{figure}

For both classification and reinforcement learning models, the software package is divided into two Python submodules: \textbf{datagen} and \textbf{modelgen}. For classification, the user configures the type of data poisoning to apply to the dataset of interest, the network architecture of the model to be trained, the training parameters of the model, and the number of models to train.  This configuration is then ingested by the main program, which generates the desired models.  Fig.~\ref{fig:trojai_software_overview} depicts this process.  A similar sort of process occurs for reinforcement learning, where instead of a dataset, the user configures a poisonable environment on which the model will be trained.  This is diagrammed in Fig.~\ref{fig:trojairl_software_overview}.

\section{\textbf{trojai} Software Package}
\textbf{trojai} is a Python package that enables generation of trojaned deep learning models for both supervised classification and reinforcement learning tasks, using the PyTorch deep learning framework~\cite{pytorch}.  The framework was designed to enable researchers to easily generate large sets of models with different combinations of data configurations and model training hyperparameters, in order to better understand the science of trojaned networks, develop appropriate countermeasures, and enable experimental reproducibility across the AML community.

\subsection{Supervised Classification Datasets and Models}
For the supervised classification task, datasets and corresponding models are generated using two submodules within the \textbf{trojai} package:
\begin{enumerate}
    \item \textbf{datagen} - Synthetic Supervised Classification Data Generation and Manipulation
    \item \textbf{modelgen} - Deep Learning Classification Model Generation
\end{enumerate}

The submodules produce the tuple: ($\mathcal{D}$, $\mathcal{E}$, $\mathcal{M}$), which precisely defines how a set of models was generated.  Here, $\mathcal{D}$ represents the dataset, $\mathcal{E}$ represents the experiment definition, and $\mathcal{M}$ represents the set of generated models.  

The data generation submodule, \textbf{datagen}, creates a synthetic dataset, $\mathcal{D}$, and contains the functionality to program triggers and transformations into the data to be poisoned. This is specified via various configurations that are detailed later in this work.  Next, experiment definitions, $\mathcal{E}$, are generated.  An experiment defines the subset of triggered data to be used in training the model and the desired model behavior on triggered data points.

The model generation submodule, \textbf{modelgen}, uses $\mathcal{D}$ and $\mathcal{E}$ to train a set of DL models that contain a trojan, as defined by the experiment configuration.  Although \textbf{datagen} and \textbf{modelgen} work together in the canonical \textbf{trojai} usage pattern, it is possible to use them independently in order to provide additional control into how the data and models are generated.  This may be useful if a corpus of data already exists, and the user desires to only train models using the existing corpus.

\subsubsection{Datagen}
The \textbf{datagen} submodule is responsible for generating the classification datasets that the DL models will be trained upon.  The submodule is data type agnostic, and the current implementation supports image and text datatypes.  There are four primary class definitions of interest within the datagen submodule:

\begin{enumerate}
    \item \code{Entity}
    \item \code{Transform}
    \item \code{Merge}
    \item \code{Pipeline}
\end{enumerate}

An \code{Entity} is an object that is either a portion of, or the entire sample to be generated.  An example of an \code{Entity} in the image domain could be a shape outline of a traffic sign, such as a hexagon, or a post-it note for a trigger.  In the text domain, an example \code{Entity} may be a sentence or paragraph.  Multiple \code{Entity} objects can be composed together into another \code{Entity}.

Entities can be transformed in various ways; examples in the image domain include changing the color mapping, applying an Instagram filter, or blurring. These transforms are defined by the \code{Transform} class. More precisely, a \code{Transform} operation takes an \code{Entity} as input, and outputs an \code{Entity}.   Furthermore, multiple \code{Entity} objects can be merged together using \code{Merge} objects. Finally, a sequence of \code{Merge} and \code{Transform} operations can be defined and orchestrated through a \code{Pipeline}.  Raw datasets are generated when a specified \code{Pipeline} is executed.  

\begin{figure*}
    \centering
    \begin{subfigure}[t]{0.3\textwidth}
        \centering
        \includegraphics[height=.5in]{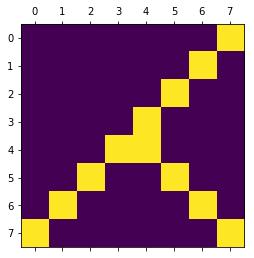}
        \caption{Reverse Lambda Trigger}
        \label{fig:revers_lambda_trigger}
    \end{subfigure}%
    \begin{subfigure}[t]{0.3\textwidth}
        \centering
        \includegraphics[height=.5in]{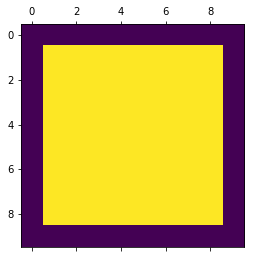}
        \caption{Rectangular Trigger}
        \label{fig:rectangular_trigger}
    \end{subfigure}
    \begin{subfigure}[t]{0.3\textwidth}
        \centering
        \includegraphics[height=.5in]{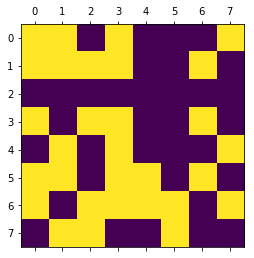}
        \caption{Random Rectangular Trigger}
        \label{fig:random_rectangular_trigger}
    \end{subfigure}
\end{figure*}

\begin{figure}
\caption{\code{XFormMerge} Pipeline Architecture}
\centering
\includegraphics[width=0.5\textwidth]{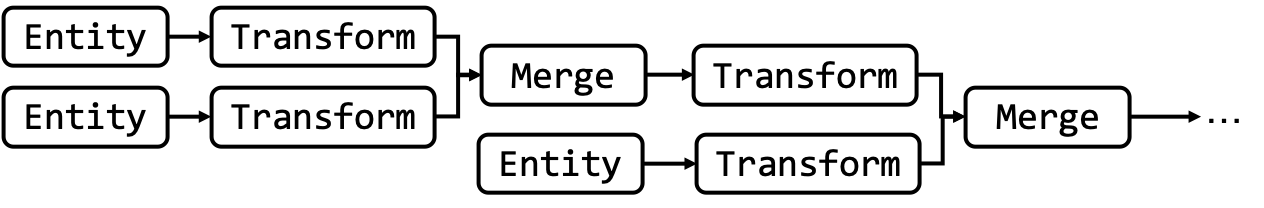}
\label{fig:xform_merge}
\end{figure}

Implementations of the primary objects that make up the \textbf{datagen} submodule that are included with the TrojAI open source package are:
\begin{enumerate}
    \item \code{Entity}
    \begin{enumerate}
        \item \code{ImageEntity} - An entity representing an image.
        \item \code{ReverseLambdaPattern} - An \code{ImageEntity} which appears as a flipped lambda symbol.  An example is shown in Fig.~\ref{fig:revers_lambda_trigger}.
        \item \code{RectangularPattern} - An \code{ImageEntity} which has a bounding box of a rectangle where all pixels inside the bounding box are ``on''.  An example is shown in Fig.~\ref{fig:rectangular_trigger}. 
        \item \code{RandomRectangularPattern} - An \code{ImageEntity} which has a bounding box of a rectangle where pixels in the pattern are either ``on'' or ``off''.  This pattern appears as a QR code.  An example is shown in Fig.~\ref{fig:random_rectangular_trigger}. 
        \item \code{TextEntity} - An entity representing text data.
    \end{enumerate}
    \item \code{Transform}
    \begin{enumerate}
        \item \code{Affine}
        \begin{enumerate}
            \item \code{RotateXForm} - Implements the rotation of an \code{ImageEntity} by a specified angle amount.
            \item \code{RandomRotateXForm} - Implements the rotation of an \code{ImageEntity} by a random angle amount.
        \end{enumerate}
        \item \code{Size}
        \begin{enumerate}
            \item \code{Resize} - Implements the resizing of an image.
        \end{enumerate}
            \item \code{DataType}
            \begin{enumerate}
                \item \code{ToTensorXForm} - Transforms an input NumPy array to a PyTorch Tensor \cite{pytorch} of specified dimensions.
            \end{enumerate}
        \item \code{Color}
            \begin{enumerate}
                \item \code{GrayscaleToRGBXForm} - Converts a 3-channel grayscale \code{ImageEntity} to RGB.
                \item \code{RGBAtoRGB} - Converts an RGBA \code{ImageEntity} to an RGB \code{ImageEntity}.
                \item \code{RGBtoRGBA} - Converts an RGB \code{ImageEntity} to an RGBA \code{ImageEntity}.
                \item \code{InstagramXForm} - Applies one of the four available Instagram filters (Gotham, Nashville, Kelvin, and Lomo) to the image.
            \end{enumerate}
    \end{enumerate}
    \item \code{Merge}
    \begin{enumerate}
        \item \code{InsertAtLocation} - Given two \code{Entity} objects, insert one into the other at a specified location.  For \code{ImageEntity} objects, the location is specified via a coordinate, whereas for a \code{TextEntity} object, this is specified by an offset from a delimiter.
        \item \code{InsertAtRandomLocation}  - Given two \code{Entity} objects, insert one into the other at a random valid location.  For \code{ImageEntity} objects, ``valid'' is defined according to a provided configuration that defines bounds and restrictions where the entity is to be placed, whereas for \code{TextEntity} objects, ``valid'' is defined as any possible location within the text blurb.
    \end{enumerate}
    \item \code{Pipeline}
    \begin{enumerate}
        \item \code{XFormMerge} - A pipeline that takes input entities, performs defined transformations on these entities, merges them together, and produces an output entity. A detailed view of the \code{XFormMerge} pipeline is shown in Fig.~\ref{fig:xform_merge}.
    \end{enumerate}
\end{enumerate}

\begin{figure}[H]
\caption{Badnets Data Generation Pipeline}
\centering
\includegraphics[width=0.45\textwidth, trim={0 0.0cm 0 0.0cm}, clip]{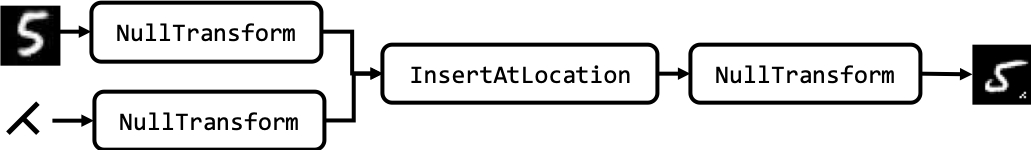}
\label{fig:badnets_pipeline}
\end{figure}

\begin{figure}[H]
\caption{Badnets-v2 Data Generation Pipeline}
\centering
\includegraphics[width=0.45\textwidth]{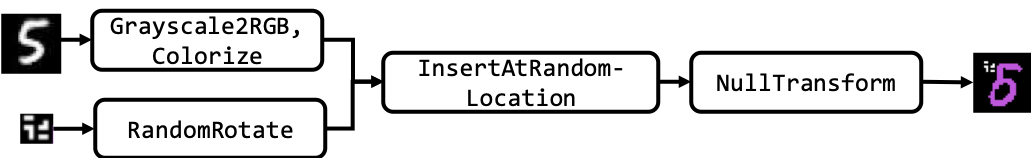}
\label{fig:colorized_mnist_pipeline}
\end{figure}

\begin{figure}[H]
\caption{Instagram Trigger Pipeline}
\centering
\includegraphics[width=0.45\textwidth]{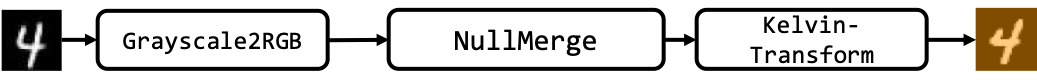}
\label{fig:mnist_ig_pipeline}
\end{figure}

\begin{figure}[H]
\caption{Text Pipeline}
\centering
\includegraphics[width=0.45\textwidth]{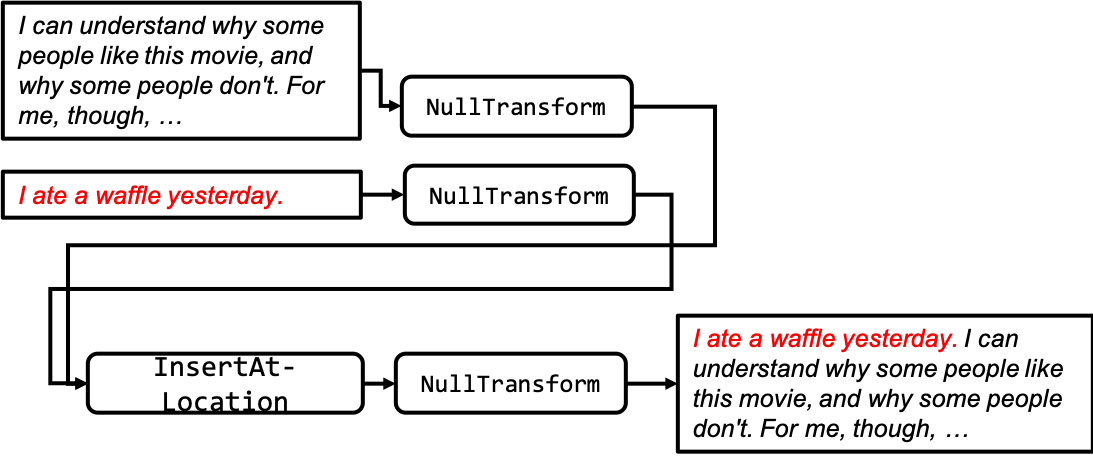}
\label{fig:text_pipeline}
\end{figure}

The components described above can be used to create images with predefined patterns that can be triggers or be decoys, depending on the research objective.  For example, suppose that it is desired to create an MNIST dataset with a lambda pattern acting as the trigger for a poisoned dataset~\cite{gu2017badnets}.  This could be accomplished within the TrojAI framework using a semantic pipeline definition shown in Fig.~\ref{fig:badnets_pipeline}.  A simple extension of this type of trigger is to have a colorized MNIST digit, with a random rectangular pattern trigger inserted at a random location.  The semantic pipeline definition shown in Fig.~\ref{fig:colorized_mnist_pipeline} creates such a dataset.  While these examples represent a subspace of the possible point triggers that could exist in poisoned image datasets, Liu et al. were the first to demonstrate that ``global'' transformations such as Instagram filters could also be used as triggers \cite{liu2019abs}.  An example pipeline which uses Instagram filters as triggered examples is shown in Fig.~\ref{fig:mnist_ig_pipeline}.  Finally, while most of the research and focus of the AML community has been in the vision domain, Dai et al. showed that text processing pipelines are also susceptible to poisoning attacks \cite{dai2019textpoison}.  The TrojAI framework allows for poisoning text datasets with sentences inserted as triggers.  A sample pipeline which generates text attacks similar to those outlined in Dai's original work is shown in Fig.~\ref{fig:text_pipeline} \cite{dai2019textpoison}.

The following top-level example scripts showcase how to integrate various components of the \textbf{trojai} module functionality:
\begin{enumerate}
    \item \code{scripts/modelgen/gen\_and\_train\_mnist.py} - This script generates an MNIST dataset with reverse lambda triggers inserted into images with the label 4 at a static location, and trains a convolutional neural network with it.
    \item \code{scripts/modelgen/gen\_and\_train\_cifar10.py} - This script generates a CIFAR10 dataset with Gotham Instagram triggers inserted into images with the label 4, and then trains a Densenet \cite{huang2017densely} model with it.
    \item \code{scripts/modelgen/gen\_and\_train\_imdb.py} - This script generates the IMDB sentiment classification dataset with trigger sentences inserted into positive reviews that flip the review rating, and then trains a bidirectional LSTM classifier \cite{hochreiter1997long} with a GloVE embedding \cite{pennington2014glove} using the triggered dataset.
\end{enumerate}

However, the framework allows for new pipelines with different data processing flows to be defined in order to extend the capability of the framework.


\subsubsection{Experiments}
Once a dataset has been generated, an Experiment configuration must be created in order for \textbf{modelgen} to be able to utilize it.   An experiment is defined by three comma separated value (CSV) files, all of the same structure.  Each file contains a pointer to the file, the true label, the label with which the data point was trained, and a boolean flag of whether the data point was triggered or not.  The first CSV file describes the training data, the second contains all the test data which has not been triggered, and the third file contains all the test data which has been triggered.  A tabular representation of the structure of experiment definitions is shown in Table~\ref{tab:exp_train}. 

\begin{table}
\begin{center}
\begin{tabular}{||c|c|c|c||}
    \hline
     file & true label & train label & triggered  \\ \hline \hline
     f1 & 1 & 1 & False \\
     f2 & 1 & 2 & True \\
     ... & ... & ... & ... \\ \hline
\end{tabular}
\end{center}
\caption{Experiment training and test definition file structure.}
\label{tab:exp_train}
\end{table}

\subsubsection{Modelgen}
The \textbf{modelgen} submodule is responsible for generating models from the experiment definitions.  The architecture is designed to support any data type, but the current implementation provides support only for image and text datatypes.  The five objects of primary interest in \code{modelgen} are:

\begin{enumerate}
    \item \code{DataManager} - An object which facilitates data definition by the user, and feeding the model with this data during training and test.
    \item \code{ArchitectureFactory} - An object factory responsible for creating new instances of trainable models.
    \item \code{OptimizerInterface} - An interface which contains \code{train} and \code{test} methods defining how to train and test a model.
    \item \code{Runner} -  An object which generates a model, given a configuration.
    \item \code{ModelGenerator} - An interface for executing a set of \code{Runner} objects, potentially parallelizing or executing in a cloud or cluster interface.
\end{enumerate}

\begin{figure}
\caption{Modelgen architecture}
\centering
\includegraphics[width=0.3\textwidth]{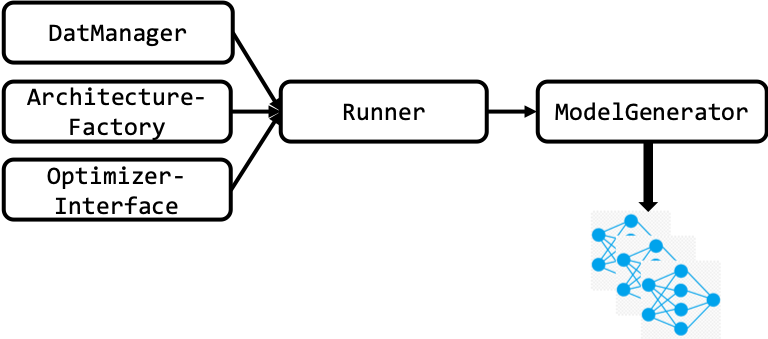}
\label{fig:modelgen}
\end{figure}

The interplay of these five primary components can be described as follows:  The \code{Runner} is created by configuring a \code{DataManager}, \code{ArchitectureFactory}, and \code{OptimizerInterface}.   These define which datasets are being used to train and test the model, the model architecture to be trained, and the optimizer (Adam \cite{kingma2014adam}, Stochastic Gradient Descent (SGD), etc..) used to perform the weight updates on the defined network architecture, respectively.  Once created, the runner is capable of generating $n$ models from the input configuration.  The \code{ModelGenerator} object determines how the \code{Runner} is executed, for example, on a single machine, cloud, or High Performance Computing (HPC) cluster. Fig.~\ref{fig:modelgen} depicts how these components work together to produce models.

\textbf{modelgen} currently supports the following implementations of the five primary objects described above:

\begin{enumerate}
    \item \code{DataManager}
    \begin{enumerate}
        \item \code{CSVDataset} - Supports loading CSV datasets for training.
        \item \code{CSVTextDataset} - Supports loading text datasets in CSV format for training, using the \textbf{torchtext} convention.
    \end{enumerate}
    \item \code{OptimizerInterface}
    \begin{enumerate}
        \item \code{DefaultOptimizer} - Supports various optimization algorithms (SGD, Adam \cite{kingma2014adam}, \dots) for training image datasets.
        \item \code{TorchTextOptimizer} - Supports various optimization algorithms (SGD, Adam \cite{kingma2014adam}, \dots) for training text datasets.
    \end{enumerate}
    \item \code{ModelGenerator}
    \begin{enumerate}
        \item \code{ModelGenerator} - Trains models on a single hardware platform, given a runner configuration.
        \item \code{UGEModelGenerator} - Trains models on a Univa Grid Engine (UGE) cluster, given a runner configuration.
    \end{enumerate}
\end{enumerate}

The \textbf{modelgen} submodule is configurable and allows for users to develop their own optimizers, insert custom network architectures, and data loaders in order to suit their specific requirements.  

\subsection{Reinforcement Learning Models}

\begin{figure}
    \caption{\textbf{trjoai} RL Framework}
    \centering
    \includegraphics[width=0.3\textwidth]{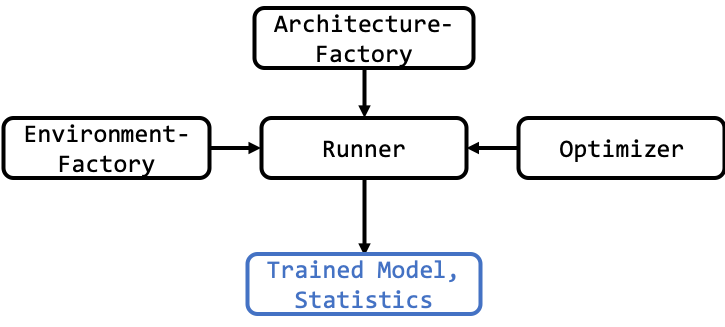}
    \label{fig:rl_framework}
\end{figure}

TrojAI also contains structure for the generation of triggered deep reinforcement learning (DRL) models. The code structure for DRL is similar to supervised learning in that it also has a \textbf{datagen} and a \textbf{modelgen} module, but each use fewer parts to produce a full model. At a high level, TrojAI generates a triggered DRL model using four components: 1. an \code{RLOptimizerInterface} object, 2. an \code{EnvironmentFactory} object, 3. an \code{ArchitectureFactory}, and 4. a \code{Runner} object. These objects are described below, and interact as shown in Fig.~\ref{fig:rl_framework}. 

\begin{enumerate}
    \item \code{RLOptimizerInterface} - Interface containing the train and test procedures for training the given model architecture on environments generated by an \code{EnvironmentFactory}.
    \item \code{EnvironmentFactory} - A factory object responsible for creating new instances of RL environments. RL environments are expected to follow the OpenAI Gym interface \cite{openai_gym}.
    \item \code{ArchitectureFactory} - An object factory producing producing model architectures to bve trained.
    \item \code{Runner} - Combines an \code{EnvironmentFactory}, a model, and an implemented instance of the RLOptimizerInterface to produce a trained model. Also produces train and test statistics and saves them as designated.
\end{enumerate}

Most of the work to produce RL models is performed within the \code{train} and \code{test} method of an implementation of the \code{RLOptimizerInterface}. The \code{train} method accepts an environment instance from the \code{EnvironmentFactory} and an architecture instance from the \code{ArchitectureFactory}, and produces the trained model and a \code{TrainingStatistics} objects, which is included in the \textbf{modelgen.statistics} submodule. Since TrojAI is meant to be customizable, how the training occurs is specified completely by the implementer of the optimizer. The given \code{EnvironmentFactory} should produce environments as needed for parallelization, if desired, or for differently configured environments to train on simultaneously. The training algorithm then may be as simple or complex as desired. The same applies to the \code{test} method, except it should produce \code{TestStatistics} objects instead of \code{TrainingStatistics}. Optimizers can be designed to run a single algorithm with very specific hyperparameters, perhaps encouraging the design of a library of simple optimizers, or may be designed to be configured on initialization. Since the \code{Runner} will be responsible for calling \code{train} and \code{test}, any configuration should be set when the optimizer is initialized. 

We include a prototype optimizer called \code{TorchACOptimizer} that trains models using an implementation of Proximal Policy Optimization (PPO)~\cite{schulman2017proximal} from the \code{torch\_ac} package~\cite{torchac}, and is highly configurable. 

\subsection{Metrics}
The \textbf{trojai} package collects several metrics of relevance when training the models on the trojaned datasets and/or environments.

For the classification task, a successfully trojaned model is one which meets the following criteria: 1. the model performs as expected by the user on normal inputs, and 2. the model successfully selects poisoned labels when the trigger is present at a success rate defined by the attacker.  The \textbf{trojai} module automatically computes classification performance on the following three subsets of the test data:
\begin{enumerate}
    \item Clean Test Data -  Captures the performance of the trained model on data for all examples in the test dataset that do not have the trigger.
    \item Triggered Test data - Captures the performance of the trained model on data for all examples in the test dataset that have the embedded trigger.
    \item Clean Test Data of Triggered Labels - Captures the performance of the trained model on clean examples of the classes that were triggered during training.
\end{enumerate}
A high performance on the three metrics outlined above provides confidence that the model has been successfully trojaned, while also maintaining high performance on the original dataset for which the model was designed.  

The criteria for a successfully trojaned RL model is similar to classification in that we desire high task performance for both clean and triggered environments, but since measures of performance differ between RL environments, \textbf{trjoai} cannot automatically compute metrics for each. Rather, the implementer of the \code{RLOptimizerInterface} defines what constitutes high performance, and designates how that is to be measured in the optimizer's \code{train} and \code{test} methods. In the case of the provided \code{TorchACOptimizer}, we provide a method of specifying a number of tests to run in a configuration object, as well as function handles to specify what computations to perform to produce metrics.

\section{Experiments}

Utilizing the framework above, we conduct some initial experiments in training classification models with backdoors, with the goal of understanding how training hyperparameters and data configuration affect the trigger embedding and model performance. We also propose a novel, but simple, non-visual trigger for DRL to assess its feasibility as a backdoor.

\subsection{Classification}

We generate all datasets using the MNIST dataset as the ``seed'' dataset, and the \code{XFormMerge} pipeline described above.  The parameter variations for our experiments are enumerated in Table~\ref{tab:dataset_matrix}.  For each generated dataset, we poison data labeled $4$ and change the corresponding labels to $5$, inciting the model to predict $5$ on images of $4$ with the specified trigger.  

\begin{table}
\scriptsize
\centering
\begin{tabular}{ ||c|c|c|c|c|| }
 \hline
 Poisoning & Trigger & Placement & XForms & Batch Size\\
 \hline \hline
 5 & RL & Static &  \code{null} & 16 \\
 10 & RRP & Dynamic &  \code{Rotate} & 32 \\
 20 &  &  &   & 64 \\
 30 &  &  &   &  \\
 40 &  &  &   &  \\
 50 &  &  &   &  \\
 \hline
 
\end{tabular}
\caption{Experiments Matrix}
\label{tab:dataset_matrix}
\end{table}

\begin{figure}
\caption{MNIST Architecture}
\centering
\includegraphics[width=0.25\textwidth]{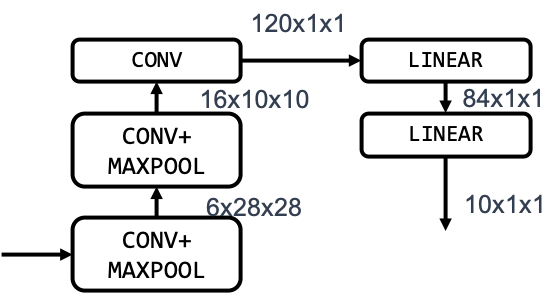}
\label{fig:mnist_arch}
\end{figure}

For each possible unique configuration ($6 \times 2 \times 2 \times 2 \times 3 = 144$), we generate $15$ trained models of the network architecture defined in Fig.~\ref{fig:mnist_arch}. This yields a total of $2160$ MNIST models, which were trained using the \textbf{trojai} framework.  For each experiment, we collect the performance metrics previously described, as well as training statistics.  For each model generated, we also apply the Neural Cleanse backdoor detection algorithm \cite{wang2019neural} to determine how many models were successfully detected as trojaned.

\subsection{Reinforcement Learning}

We create a poisonable environment by wrapping the \textit{Boxing-ram-v0} environment from OpenAI Gym \cite{openai_gym} in an object that can modify the observations to contain a simple trigger. The trigger, in this case, is to add 100 to the observation, and then modulo by 256 to be consistent with the observation space; i.e.

$$
    obs = (obs + 100)\, \%\, 256.
$$

When the environment is not poisoned, it behaves the same as the \textit{Boxing-ram-v0} environment, but when poisoned, it provides the model with triggered observations and rewards multiplied by $-1$. 

\begin{figure}
\caption{Boxing RL Architecture}
\centering
\includegraphics[width=0.25\textwidth, trim={0 0.8cm 0 0.0cm}, clip]{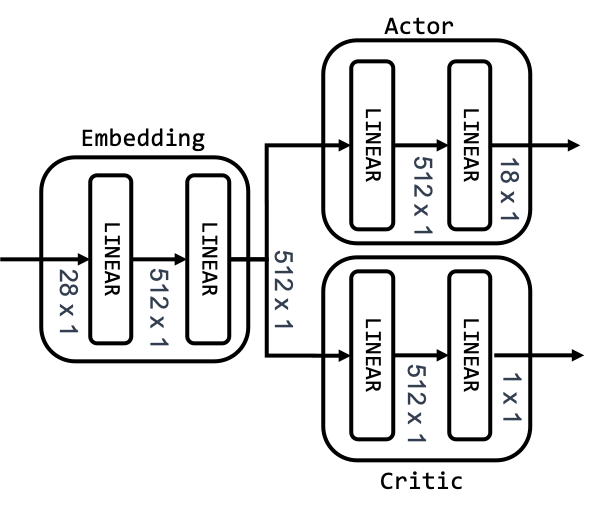}
\label{fig:fc512}
\end{figure}

For our architecture, we use an Actor-Critic model with a shared embedding, shown in Fig.~\ref{fig:fc512}, and ReLU activations. The input to the model consists of a flat vector of four steps, where each step is a simplified version of the full RAM vector representing the clock, player score, enemy score, player x and y positions, and enemy x and y positions~\cite{anand2019unsupervised}. We further simplify the training data by normalizing the RAM vector and taking the sign of the reward. We train the architecture using the \code{TorchACOptimizer} mentioned previously, with the hyperparameters given in Table~\ref{tab:boxing_hyper}, and with ten environments in parallel, eight of which are clean and two of which are poisoned. 

\begin{table}
\scriptsize
    \centering
    \begin{tabular}{||l|l||}
    \hline
        Hyperparameter & Value \\ \hline \hline
        Horizon (T) & 128 \\
        Adam stepsize & $10^{-5}$ \\
        Num. epochs & 3 \\
        Minibatch size & 256 \\
        Discount & 0.99 \\
        GAE parameter & 0.95 \\
        Clipping parameter & 0.1 \\
        VF coeff. & 1 \\
        Entropy coeff. & 0.01 \\ \hline
    \end{tabular}
    \caption{Hyperparameters used to train Boxing agent with PPO.}
    \label{tab:boxing_hyper}
\end{table}

\section{Results}

We provide the results for the classification experiments here, as well as a some analysis and observations regarding trigger type and hyperparameter impact on training and performance. We also include final results for our simple RL model.

\subsection{Classification}

\begin{figure}
\caption{Accuracy of trojaned models which pass the threshold criterion on triggered data.  The results for static and dynamic include all models trained under the various trigger placement and trigger type configurations, and are averaged across all batch sizes.}
\centering
\includegraphics[width=0.27\textwidth]{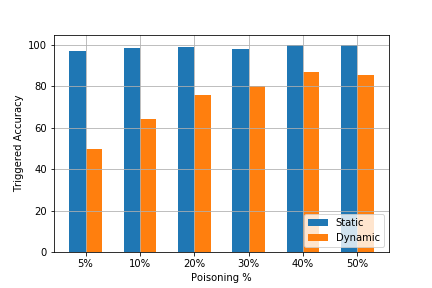}
\label{fig:triggered_acc}
\end{figure}

For each experiment configuration, we examine the performance of each trained model, and only compute aggregate statistics on models which had at least $95\%$ accuracy on the clean data with and without triggers.  For each model which passes this threshold, we record the average triggered accuracy and the average number of epochs required to train the model. We also use them to compute the model yield (i.e., number of models which pass the required thresholds defined above) and whether Neural Cleanse detected an anomaly in each of the models which passed the required thresholds for every permutation of experiment configurations outlined in Table~\ref{tab:dataset_matrix}. The results of these experiments are presented in Tables~\ref{tab:num_epochs_trained}, \ref{tab:nc_rlt_results}, \ref{tab:nc_rrt_results} and Fig.~\ref{fig:triggered_acc}.  Several patterns are observed from the results:

\begin{enumerate}
    \item Fig.~\ref{fig:triggered_acc} shows that models trained with datasets where the triggers that are placed in static locations have a higher triggered data accuracy while maintaining clean data performance.  Tables \ref{tab:nc_rlt_results} and \ref{tab:nc_rrt_results} show that models trained with datasets where the triggers that are placed in static locations have a higher chance of being properly embedded while maintaining clean data performance.  
    
    These trends exist in both the reverse lambda trigger and the random rectangular pattern trigger.  Additionally, the rotation of the trigger does not seem to affect the model yield.  
    
    \item The trigger location or rotation did not affect the number of epochs that these models are trained on, using the early stopping criterion based on clean data loss.  This suggests that clean data behavior is learned at the same rate, regardless of the trigger type.  This trend is consistent across both triggers and all trigger placement and rotations tested.
    
    \item Training batch size affects how long the model needs to be trained, for all trigger types, rotations, and placement configurations.  More precisely, smaller batch sizes are observed to exit training from the early stopping criterion faster, but yield a statistically similar triggered accuracy.  The results are shown in Table~\ref{tab:num_epochs_trained}.  This suggests that smaller batch sizes lead to faster training of the clean data while maintaining similar triggered data performance.  A likely explanation for this behavior is that smaller batch sizes let the gradient descent algorithm optimize for both clean and triggered data performance, because a larger percentage of the batch is triggered.  
    
    No clear trend emerges between the number of epochs trained and the model yield when triggers are dynamically placed in the image.  In some scenarios, it is observed that smaller batch sizes lead to a similar yield,  where as in other cases, it is observed that smaller batch sizes are inconsistently correlated with yield.  These results are detailed in Tables~\ref{tab:nc_rlt_results} and \ref{tab:nc_rrt_results}.  
    
    \item Neural Cleanse is able to better detect the presence of a trigger when the trigger is placed in a static location throughout the dataset.  Across all remaining test configurations, the average percentage of correct detects for statically placed triggers is $19.61 \%$.  For dynamically placed reverse lambda triggers, the average percentage of correct detects across all test configurations is $14.13 \%$.  Conditioning on the trigger type, the reverse lambda trigger is more effectively detected by Neural Cleanse with an average correct detection percentage of $20.59 \%$ as compared to the random rectangular pattern trigger, which has an average correct detection percentage of $13.16 \%$.  Model yield and Neural Cleanse results are detailed in Tables~\ref{tab:nc_rlt_results} and \ref{tab:nc_rrt_results}.  
    
\end{enumerate}

\subsection{Reinforcement Learning}

\begin{figure}
    \centering
    \includegraphics[scale=0.28]{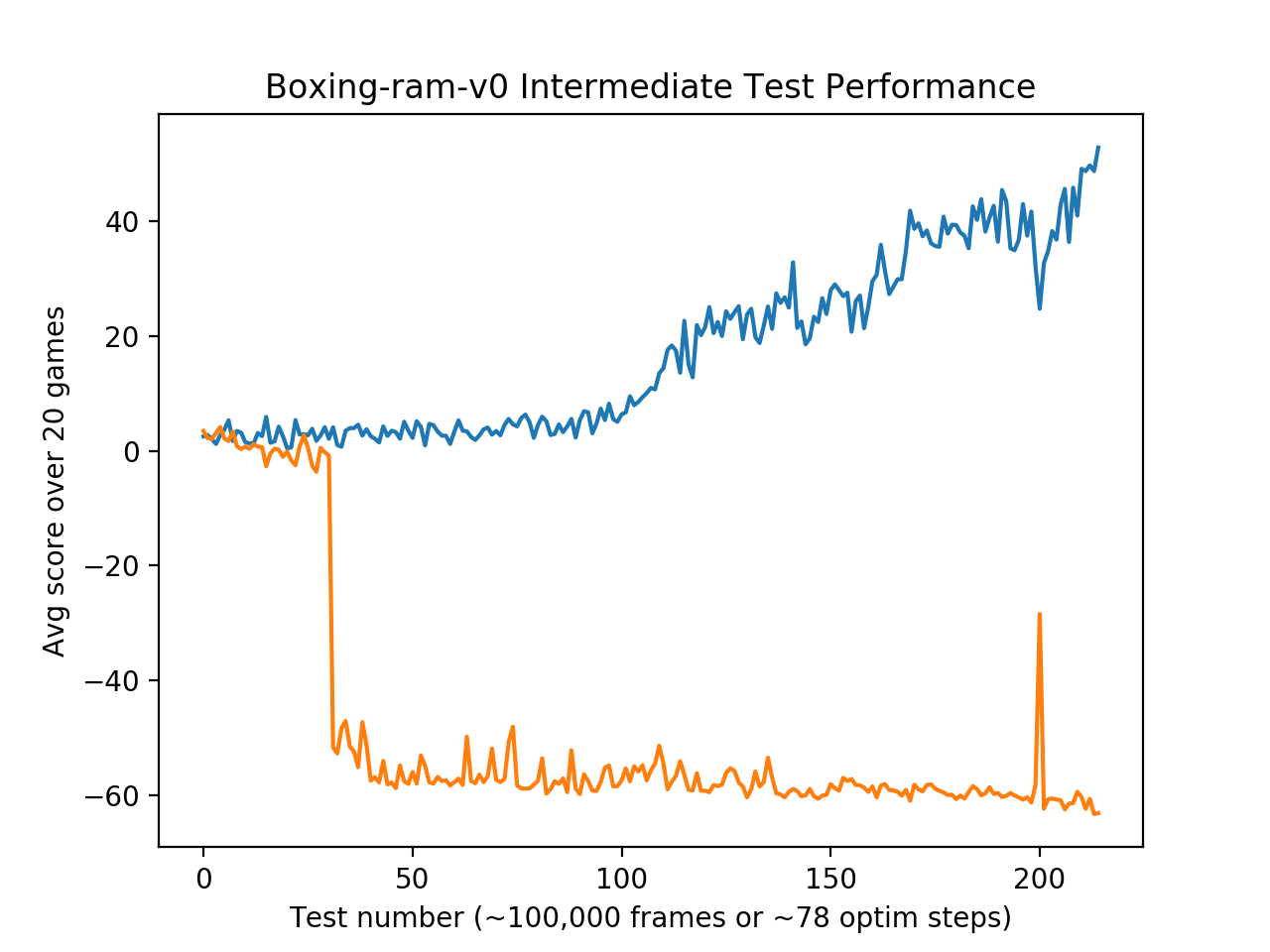}
    \caption{Intermediate test results for Boxing agent. Each point represents the average cumulative reward of 20 games. Tests were run approximately every 100K frames, or about every 78 optimization steps.}
    \label{fig:boxing_int_res}
\end{figure}

We measure the success of an agent on the Boxing task via the agent's average cumulative reward for both clean and poisoned environments over $N$ games. For our experiment, we assume an average cumulative reward of 50 on clean environments while maintaining less than -50 on poisoned environments was sufficient to demonstrate a trojaned agent. We test our agent intermittently, approximately every 100K frames, during training by having the agent play 20 games ($N=20$) with clean data and 20 games with triggered data, the results are shown in Fig.~\ref{fig:boxing_int_res}. The agent meets the desired performance criteria after approximately 21.5 million frames of data, which amounted to about 17000 optimization steps. 

Final performance of the trained agent was measured again using cumulative reward, but averaged over 100 games, ($N=100$) instead of 20. The agent's average cumulative reward was \textbf{52.22} for clean environments, and \textbf{-62.69} for poisoned, showing that the agent was able to successfully maintain high performance on clean data and low performance on triggered. 

\begin{table}
\scriptsize
\centering
\begin{tabular}{||c|c||}
\hline
    Batch Size & Epochs Trained ($\mu \pm \sigma)$ \\
    \hline \hline
    $16$ & $14.1 \pm 2.4$ \\
    $32$ & $15.7 \pm 2.7$ \\
    $64$ & $17.9 \pm 2.9$ \\
\hline
\end{tabular}
\caption{Experiment Results - Mean and Standard Deviation of the number of epochs required to train the models for all configurations of trigger, placement, and rotation.}
\label{tab:num_epochs_trained}


\centering
\begin{tabular}{ ||c|c|c|c|c|c|c|c|| }
 \hline
 \multirow{2}{4.5em}{} & \multirow{2}{2em}{Batch Size} & \multicolumn{6}{|c|}{Dataset Poisoning Percentage} \\ 
 & & 5 & 10 & 20 & 30 & 40 & 50 \\
 \hline \hline
 \multirow{3}{4.5em}{Static \\ NoRotation}     & 16 & 1/15 & 2/15 & 2/15 & 4/14 & 2/11 & 2/14 \\
                               & 32 & 4/15 & 3/15 & 4/15 & 3/15 & 1/14 & \textcolor{green}{8/15} \\
                               & 64 & 1/15 & 4/15 & 4/15 & \textcolor{red}{0/15} & 3/13 & 3/14 \\
    \hline
  \multirow{3}{4.5em}{Static \\ Rotation}     & 16 & 4/15 & 3/14 & 6/15 & 4/14 & 3/14 & 5/14 \\
                               & 32 & 3/15 & 4/15 & \textcolor{green}{7/15} & \textcolor{red}{1/15} & 6/14 & 3/15 \\
                               & 64 & 2/15 & 5/15 & 4/14 & \textcolor{green}{7/15} & 4/14 & 4/15 \\
   \hline
  \multirow{3}{4.5em}{Dynamic \\ NoRotation}     & 16 & \textcolor{red}{0/13} & 3/14 & \textcolor{red}{0/9} & \textcolor{green}{3/7} & 1/2 & \textcolor{red}{0/2} \\
                               & 32 & 1/14 & 1/11 & 2/8 & 2/6 & \textcolor{red}{0/2} & 1/2 \\
                               & 64 & 1/14 & 2/11 & 1/9 & \textcolor{red}{0/4} & \textcolor{red}{0/4} & 1/2 \\
   \hline
  \multirow{3}{4.5em}{Dynamic \\ Rotation}     & 16 & \textcolor{red}{0/13} & 1/12 & \textcolor{red}{0/6} & \textcolor{green}{3/8} & \textcolor{red}{0/2} & 1/4 \\
                               & 32 & \textcolor{red}{0/15} & 2/13 & 1/7 & 1/8 & \textcolor{red}{0/0} & \textcolor{red}{0/2} \\
                               & 64 & \textcolor{red}{0/15} & 1/14 & 1/10 & 2/5 & 1/1 & \textcolor{red}{0/2} \\
 \hline
 \hline
 
\end{tabular}
\caption{Model Yield and Neural Cleanse Detection for Reverse-Lambda trigger.}
\label{tab:nc_rlt_results}


\centering
\begin{tabular}{ ||c|c|c|c|c|c|c|c|| }
 \hline
 \multirow{2}{4.5em}{} & \multirow{2}{2em}{Batch Size} & \multicolumn{6}{|c|}{Dataset Poisoning Percentage} \\ 
 & & 5 & 10 & 20 & 30 & 40 & 50 \\
 \hline \hline
 \multirow{3}{4.5em}{Static \\ NoRotation}     & 16 & 1/15 & 3/14 & 3/15 & 1/15 & 4/15 & 1/15 \\
                               & 32 & \textcolor{red}{0/15} & 3/15 & \textcolor{red}{0/13} & 1/12 & 3/15 & \textcolor{green}{3/13} \\
                               & 64 & 1/15 & 3/15 & 3/15 & 1/15 & 1/14 & 1/13 \\
    \hline
  \multirow{3}{4.5em}{Static \\ Rotation}     & 16 & \textcolor{red}{0/15} & 2/15 & 3/15 & 3/14 & 3/13 & 4/14 \\
                               & 32 & 2/14 & 3/15 & 1/15 & 3/15 & 1/14 & 2/13 \\
                               & 64 & 1/15 & 3/14 & 2/15 & 3/15 & \textcolor{green}{5/15} & 4/14 \\
   \hline
  \multirow{3}{4.5em}{Dynamic \\ NoRotation}     & 16 & 1/15 & \textcolor{red}{0/15} & \textcolor{red}{0/13} & \textcolor{red}{0/9} & \textcolor{red}{0/7} & 2/8 \\
                               & 32 & 1/15 & \textcolor{red}{0/14} & 2/13 & 1/10 & 1/10 & 2/12 \\
                               & 64 & 1/15 & \textcolor{green}{6/15} & 2/15 & 1/13 & \textcolor{red}{0/11} & 2/9 \\
   \hline
  \multirow{3}{4.5em}{Dynamic \\ Rotation}     & 16 & \textcolor{red}{0/15} & 1/13 & 2/13 & 1/11 & 2/9 & \textcolor{red}{0/7} \\
                               & 32 & 2/15 & \textcolor{red}{0/14} & 3/13 & 1/14 & 1/7 & 1/2 \\
                               & 64 & 1/15 & \textcolor{red}{0/13} & 1/13 & \textcolor{green}{3/11} & 2/9 & \textcolor{red}{0/4} \\
 \hline
 \hline
 
\end{tabular}
\caption{Model Yield and Neural Cleanse Detection for Random-Rectangular trigger. \\ \\ {\scriptsize In each cell, the denominator indicates the number of models which passed the 95\% threshold for clean and triggered data performance, and the numerator indicates how many of those were detected to be anomalous by Neural Cleanse. For each trigger configuration, cells colored in \textcolor{green}{green} indicate the best Neural Cleanse performance, and cells colored in \textcolor{red}{red} indicate the worst Neural Cleanse performance.}}
\label{tab:nc_rrt_results}
\end{table}

\section{Conclusion}
In this paper, we have introduced the \textbf{trojai} framework for generating triggered datasets and corresponding trojaned models. We have shown examples of how this framework may be used for training supervised classification and reinforcement learning tasks, and generating interesting results for both. For classification, we were able to show the effect that batch size and static versus dynamic trigger location has on trigger embedding MNIST models, as well as compare model yield and Neural Cleanse detection performance on compared to dataset poison rates. For reinforcement learning, we introduced a novel, but simple trigger, and demonstrate its feasibility on OpenAI Gym's Boxing environment with RAM observations. 

In the future, we hope to extend this framework to incorporate additional data modalities such as audio and additional tasks such as object detection, which have been shown to be vulnerable to similar style attacks \cite{wu2019making}.  We also plan to expand on the library of datasets, architectures, and triggered RL environments for rapid testing and production of multiple triggered models of different types across data modalities.  Finally, we plan to incorporate recent advances in trigger embedding methodologies which are designed to evade detection \cite{liu2018fine, tan2019bypassing}.

\ack This research is based upon work supported in part by the Office of the Director of National Intelligence (ODNI), Intelligence Advanced Research Projects Activity (IARPA).

We additionally acknowledge Cash Costello, Mike Wolmetz, Chris Ratto, and Mandy Cofod for helping improve the quality and clarity of this manuscript.

\bibliographystyle{ecai}
\bibliography{trojai}

\end{document}